\documentclass[conference]{IEEEtran}
\IEEEoverridecommandlockouts
\usepackage{graphicx}
\usepackage{amsmath}
\usepackage{cite}
\usepackage{multirow}
\usepackage[font=footnotesize]{caption}
\usepackage{hyperref} 
\usepackage{balance}
\def\BibTeX{{\rm B\kern-.05em{\sc i\kern-.025em b}\kern-.08em
    T\kern-.1667em\lower.7ex\hbox{E}\kern-.125emX}}
\begin{document}

\title{Hand Over or Place On The Table? A Study On Robotic Object Delivery When The Recipient Is Occupied}

\author{\IEEEauthorblockN{Thieu Long Phan}
\IEEEauthorblockA{Deakin University \\
Melbourne, Australia \\
paul.phan@deakin.edu.au}
\and
\IEEEauthorblockN{Akansel Cosgun}
\IEEEauthorblockA{Deakin University \\
Melbourne, Australia \\
akansel.cosgun@monash.edu.au}
}

\maketitle

\begin{abstract}
This study investigates the subjective experiences of users in two robotic object delivery methods: direct handover and table placement, when users are occupied with another task. A user study involving 15 participants engaged in a typing game revealed that table placement significantly enhances user experience compared to direct handovers, particularly in terms of satisfaction, perceived safety and intuitiveness. Additionally, handovers negatively impacted typing performance, while all participants expressed a clear preference for table placement as the delivery method. These findings highlight the advantages of table placement in scenarios requiring minimal user disruption.
\end{abstract}


\section{Introduction}
Human-robot interaction (HRI) research often focuses on controlled environments where users engage directly with robots, particularly in tasks like object handovers \cite{ortenzi2021object}. However, in reality, users are frequently distracted or multitasking, which makes the current research limited. This creates a gap in understanding how robotic systems can adapt to users who are not always paying full attention. This study seeks to address this gap by comparing two object delivery methods—handover, where the robot directly hands an object to the user, and placing objects on a table, where the user retrieves the object themselves—when users are occupied with another task. Specifically, participants were engaged in a typing game, requiring them to divide their attention between the game and the robot’s actions. We assess the impact of these delivery methods on key user experience metrics, including satisfaction, safety, intuitiveness, confidence and mental demand while users are distracted by a concurrent task. The findings aim to inform the design of context-aware robotic systems that can dynamically adjust based on the user’s attentiveness.

\section{User Study Design}
For this study, we used the Hello Robot Stretch 2, a robot platform designed for human environments, featuring a prismatic arm and a Lidar sensor for navigation, as well as a gripper at the end of the arm to hand and place objects. This setup allows the robot to perform both object delivery methods—handover and placing objects on a table—with precision. To ensure smooth and predictable delivery, we employed a simple waypoint navigation system using Robot Operating System (ROS2) that allows the robot to navigate to multiple destinations. This system guided the robot along a pre-defined path to the delivery point, maintaining consistency and minimizing distractions for the user. Once the robot arrived at the delivery destination, it would either hand the object—in this experiment, a plush toy—directly to the user or place it on the table for later retrieval, all while participants were engaged in a typing game.

The user study involved 15 participants, aged 18 to 24, selected primarily from university students. The experiment aimed to assess the effectiveness of two object delivery methods—handover and placing the object on a table—while participants were required to engage in the typing game, where they needed to type words as fast as possible with least mistakes while simultaneously interacting with the robot during object delivery. This multitasking scenario was designed to simulate real-world conditions where users may be distracted or divided in attention. The ethical approval for this experiment was granted by the Deakin University Ethics Board under ID SEBE-2024-29. Participants provided informed consent by reading and signing a handout at the start of the study.

\setlength{\tabcolsep}{0.5mm}
\begin{table}[t!]
\centering
\begin{tabular}{|c|c|}
\hline
\multicolumn{2}{|c|}{Delivery Method} \\
\hline
Handover & Placing on Table \\
\hline
\raisebox{-0.7mm}{\includegraphics[height=3cm]{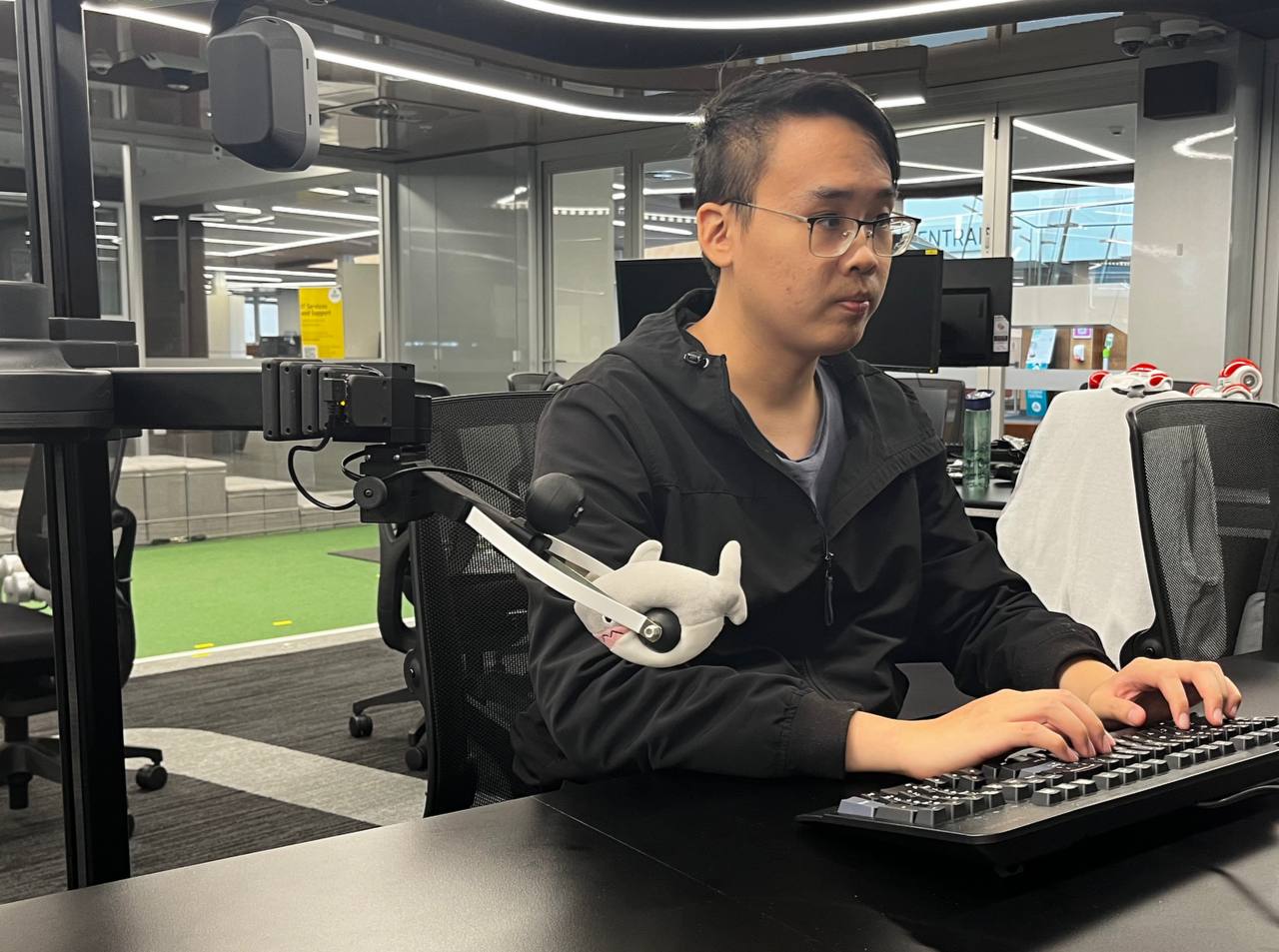}} & 
\raisebox{-0.7mm}{\includegraphics[height=3cm]{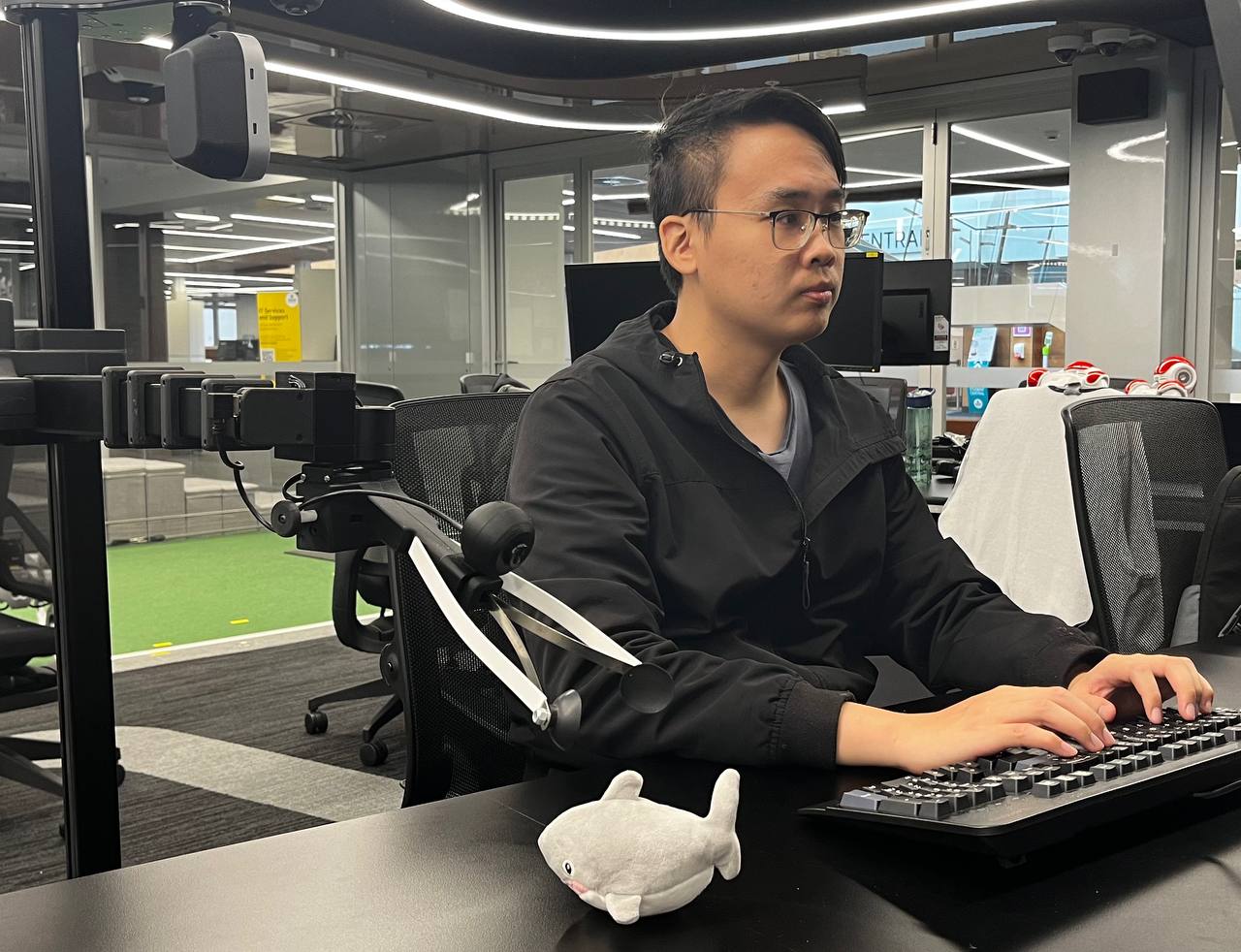}} \\
\hline
\end{tabular}
\caption{Illustration of the two object delivery methods when user is occupied: (a) Handover, where the robot directly hands the object to the user, and (b) Placing on the table, where the user retrieves the object themselves.}
\label{table:delivery_methods}
\end{table}
\setlength{\tabcolsep}{0.15cm}

Each participant experienced both delivery methods while being occupied. The order of the delivery method was counterbalanced to prevent order effects. To manage participant expectations and minimize confusion, a clearly visible label indicating the current delivery method (either handover or placing the object on the table) was placed near the participants. This helped ensure participants were aware of the delivery method in use, reducing inefficiencies caused by uncertainty, such as waiting for an action that was not happening or prematurely extending their hand.

Feedback on satisfaction, safety, intuitiveness, confidence, and mental demand was collected using a 7-point Likert scale survey after each delivery method was completed. These metrics were selected to provide a comprehensive view of user experience: satisfaction reflecting overall enjoyment, safety assessing perceived risk, intuitiveness measuring ease of interaction, confidence indicating trust in the robot, and mental demand evaluating the cognitive load of the interaction. We also measured the typing performance and asked the participants for their preferred delivery method at the conclusion of the user study. The visual comparison of the two delivery methods, shown in Table~\ref{table:delivery_methods}, illustrates how the robot interacts with participants during the typing game.

\section{Results}

The results are summarized in Table~\ref{table:occupied_state}. Participants reported significantly higher ratings for satisfaction, safety, intuitiveness, and confidence when the object was placed on the table compared to the handover method, with all p-values below the 0.05 significance threshold. All 15 participants preferred table placement in the post-study questionnaire. These results indicate a strong preference for placing the object on the table when users are multitasking. However, mental demand did not show significant differences, suggesting no additional cognitive load imposed by either delivery method, and was therefore excluded from further analysis.

\begin{table}[h!]
\centering
\normalsize
\begin{tabular}{|l|c|c|l|}
\hline
\textbf{Metric} & \textbf{Handover} & \textbf{Place on table} & \textbf{p-value} \\
\hline
Satisfaction    & 4.53              & 6.67                  & 9e-05$^{*}$          \\
Safety          & 4.60              & 6.67                  & 6e-05$^{*}$          \\
Intuitiveness   & 4.40              & 6.40                  & 6e-05$^{*}$          \\
Confidence      & 5.60              & 6.40                  & 0.017$^{*}$          \\
Mental demand      & 3.80              & 3.53                  & 0.057          \\
\hline
\end{tabular}
\caption{Mean survey scores comparing handover and place on table. Statistically significant results are indicated by $^{*}$.}
\label{table:occupied_state}
\end{table}

The preference for placing the object on the table suggests that participants found this method more intuitive and less disruptive. The ability to retrieve the object at their own pace, rather than relying on the robot to deliver it directly while they were distracted, likely contributed to the increased satisfaction and perceived safety ratings. These findings underscore the importance of user control in interactions with robots, particularly in multitasking environments where users' attention is divided. Confidence in the robot's ability was also significantly higher for the placing the object on the table method. When participants could retrieve the object at their convenience, they reported feeling more in control of the interaction. This suggests that autonomous robots that allow for user agency in interactions could help improve trust and ease of use. Regarding game performance, participants experienced a 12-19\% decrease in typing performance with the handover method, which indicates that the need to interrupt their work to receive the object from the robot negatively impacted their typing speed and accuracy. In contrast, placing the object on the table allowed participants to maintain their focus on the typing task with minimal disruption. Post-study interviews further supported these findings, with many participants noting that placing the object on the table allowed them to continue working without disruption, making the interaction feel less intrusive and more suitable for multitasking. While handover was appreciated for its immediacy, participants felt it was more challenging when engaged in another task, as the need to focus on the robot increased anxiety about interacting with the robot properly. These reflections underscore the advantage of placing the object on the table in occupied scenarios, as it enables more seamless, low-stress interactions, while handover may still be better suited for situations where immediacy is critical.

\section{Discussion \& Conclusion}

The findings of this study highlight the importance of adapting robotic delivery methods when users are occupied with other tasks. In these scenarios, placing the object on a table proved to be the preferred method, with significant advantages in satisfaction, safety, and intuitiveness. This approach minimizes disruptions and allows users to continue their primary activities without needing to directly engage with the robot. By contrast, handing the object directly showed lower user ratings, suggesting that direct object delivery is less effective when users are multitasking. From a robot design perspective, these findings suggest that if a robot is only capable of handover, it should wait until users are not occupied before attempting to deliver an object. However, if the robot can perform both handover and placing objects on the table, it should use table placement when users are engaged in another task and reserve handover for when users are fully available for interaction. In such scenarios, it would also beneficial for the robot to clearly communicate its intended delivery method to the user\cite{newbury2022visualizing}.

Our study addresses a critical gap in human-robot interaction (HRI) research, which has predominantly focused on scenarios involving fully attentive users. Existing literature on human-robot handovers has overlooked the impact of user attentiveness \cite{ortenzi2021object}. We believe that HRI scenarios should be evaluated under conditions where users are occupied with other tasks, reflecting real-world contexts more accurately. While there are studies that have explored object handovers in ideal conditions, they have not considered the effects of distraction or multitasking on user experience \cite{choi2009handover}. This study demonstrates that robotic systems must adapt delivery methods to users' attention levels, providing an important contribution to the HRI field.

In conclusion, this study suggests that robots should dynamically adjust their delivery methods based on user attentiveness. When users are occupied, robots should opt for placing the object on a table to minimize disruption and ensure a smoother interaction. This adaptive approach not only improves the user experience in real-world, multitasking environments but also enhances the robot's effectiveness in assisting users who may not be fully engaged. Future research could focus on refining these adaptive strategies, integrating real-time feedback and additional contextual factors to further optimize robotic interactions \cite{tian2023crafting}.

\bibliographystyle{plain}
\balance
\bibliography{refs} 

\end{document}